\documentclass[letterpaper, 10pt, conference]{ieeeconf} 
\IEEEoverridecommandlockouts 
\usepackage{cite}
\usepackage{amsmath,amssymb,amsfonts}
\usepackage{algorithmic}
\usepackage{graphicx}
\usepackage{textcomp}
\usepackage{xcolor}
\usepackage{booktabs}
\usepackage{siunitx}
\usepackage{comment}
\usepackage{multirow} 
\usepackage{subcaption}
\usepackage{url}
\usepackage{hyperref}
\def\BibTeX{{\rm B\kern-.05em{\sc i\kern-.025em b}\kern-.08em
    T\kern-.1667em\lower.7ex\hbox{E}\kern-.125emX}}

\title{HiMAP: History-aware Map-occupancy Prediction with Fallback}
\author{
Yiming Xu$^{1}$,
Yi Yang$^{2}$,
Hao Cheng$^{3}$,
Monika Sester$^{1}$
\thanks{$^{1}$\,Yiming Xu and Monika Sester are with the Institute of Cartography and Geoinformatics, Leibniz University Hannover, Appelstr. 9a, 30167 Hannover, Germany {\tt\small\{Yiming.Xu, Monika.Sester\}@ikg.uni-hannover.de}}
\thanks{$^{2}$\,Yi Yang is with the Institute of Information Processing, Leibniz University Hannover, Schneiderberg 32, 30167 Hannover, Germany {\tt\small yangyi@tnt.uni-hannover.de}}
\thanks{$^{3}$\,Hao Cheng is with the Faculty of Geo-Information Science and Earth Observation, University of Twente, 7522 NH Enschede, The Netherlands. {\tt\small h.cheng-2@utwente.nl}}
}
\begin{document}
\maketitle

\begin{abstract}
Accurate motion forecasting is critical for autonomous driving, yet most predictors rely on multi-object tracking (MOT) with identity association, assuming that objects are correctly and continuously tracked. When tracking fails due to, e.g., occlusion, identity switches, or missed detections, prediction quality degrades and safety risks increase. We present \textbf{HiMAP}, a tracking-free, trajectory prediction framework that remains reliable under MOT failures. HiMAP converts past detections into spatiotemporally invariant historical occupancy maps and introduces a historical query module that conditions on the current agent state to iteratively retrieve agent-specific history from unlabeled occupancy representations. The retrieved history is summarized by a temporal map embedding and, together with the final query and map context, drives a DETR-style decoder to produce multi-modal future trajectories. This design lifts identity reliance, supports streaming inference via reusable encodings, and serves as a robust fallback when tracking is unavailable. On Argoverse~2, HiMAP achieves performance comparable to tracking-based methods while operating without IDs, and it substantially outperforms strong baselines in the no-tracking setting, yielding relative gains of 11\% in FDE, 12\% in ADE, and a 4\% reduction in MR over a fine-tuned QCNet. Beyond aggregate metrics, HiMAP delivers stable forecasts for all agents simultaneously without waiting for tracking to recover, highlighting its practical value for safety-critical autonomy.
The code is available under: \url{https://github.com/XuYiMing83/HiMAP}.
\end{abstract}

\section{Introduction}
In autonomous driving and robotic navigation, accurately predicting the future trajectories of surrounding agents is critical for understanding the environment and making safe decisions. A primary motivation for trajectory prediction is that reliable estimates of nearby agents’ future motions enable automated systems to plan safer paths. However, trajectory prediction is inherently unstable, as it is embedded within a complex pipeline of sequential tasks, including detection, segmentation, tracking, trajectory forecasting, and planning~\cite{hu2023planning, ye2023fusionad}. Yet, integrating existing methods~\cite{zhou2023query, li2024bevformer, wang2023exploring} into a single system often introduces instability across the pipeline.

\begin{figure}[t!]
\centerline{\includegraphics[width=1\linewidth]{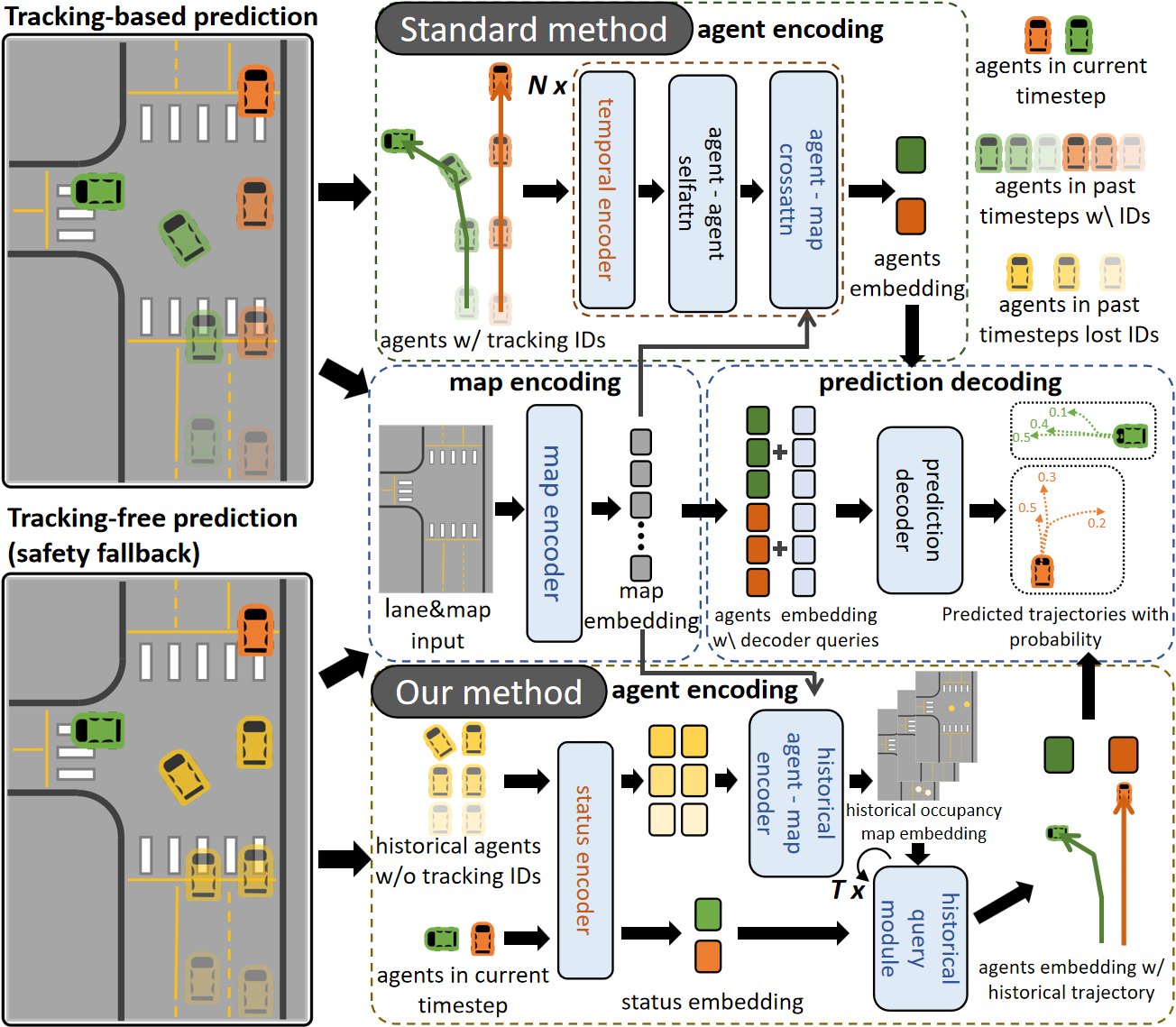}}
\caption{Comparison between tracking-based prediction and our tracking-free fallback. 
Tracking-based methods rely on stable identity association, but fail when tracking breaks. 
Our tracking-free module provides a complementary safety mechanism by reconstructing history from historical occupancy maps, ensuring reliable prediction even under tracking failures. 
Our method matches the current agent state to historical occupancy maps, implicitly recovering its past states without explicit IDs for robust trajectory prediction.}
\vspace{-18pt}
\label{fig:1}
\end{figure}

Trajectory forecasting methods heavily depend on detection and tracking modules, assuming reliable tracks within a certain range to generate multi-modal predictions of future motions. In practice, however, tracking is prone to failures such as flickering, identity drift, and ID switches, especially under occlusion or in dense multi-agent scenes~\cite{feng2024rgbt,zhang2023motrv2,liu2025sparsetrack}. These issues often lead to incomplete or erroneous histories, causing prediction accuracy and robustness to degrade.  

In this work, we introduce \textbf{HiMAP}, a trajectory prediction module designed to mitigate failures caused by unstable or missing tracking. Our key idea is to track agents implicitly by leveraging historical detections alone---without requiring consistent tracking identities. Even if the tracking module collapses entirely, HiMAP can still produce reliable trajectory forecasting based solely on cached historical detections, enhancing the robustness and safety of autonomous systems.

Our method converts past time-based trajectories into space-based occupancy maps for tracking-free training. Specifically, we first encode each agent’s state and high-definition (HD) map information (including lane geometry). At each timestep, we integrate detected agent states into the map to build a sequence of historical occupancy maps. The observation module then uses the current agent state and the full history of occupancy maps to construct a query that iteratively decodes each past frame, implicitly reconstructing the agent’s historical trajectory. Finally, multi-modal future trajectories are predicted from the decoded history, the current agent state, and the map context, as shown in Fig.~\ref{fig:1}.
The core principle is that \textit{by projecting detections as occupancy information on per-frame maps, HiMAP can recover agents’ motion history even without explicit track IDs}. By aggregating these observations over time, the module stabilizes trajectory predictions and safeguards downstream decision-making in safety-critical systems.

Our main contributions are summarized as follows:
\begin{itemize}
    \item We propose a trajectory prediction framework that remains reliable when tracking is unstable or unavailable. Without requiring tracking identities, our method reconstructs historical states from detections alone and predicts future trajectories with improved stability.
    \item We design a scene-level encoding of rich map context and a recurrent observation mechanism over historical occupancy maps. Parameters are shared across timesteps, and the module is aligned with standard agent--map encoders, enabling direct integration into existing forecasting models with minimal overhead.
    \item On the Argoverse~2 dataset, HiMAP achieves performance close to baselines that assume perfect tracking. In tracking failure scenarios, our method significantly improves robustness and substantially narrows the performance gap to tracking-based approaches. In particular, compared with tracking-based baselines in the tracking failure setting, HiMAP achieves relative improvements of 11\% in FDE, 12\% in ADE, and 4\% in MR.
\end{itemize}

\section{Related Works}
\subsection{Tracking in Motion Forecasting}
Most state-of-the-art forecasting models rely on multi-object tracking (MOT) to provide identity-consistent histories for agents. Trackers such as AB3DMOT~\cite{weng2020ab3dmot}, CenterPoint with tracking~\cite{yin2021centerpoint}, and recent detection-based MOT methods including ByteTrack~\cite{zhang2022bytetrack}, MOTRv2~\cite{zhang2023motrv2}, and BoT-SORT~\cite{aharon2022bot} are commonly adopted in autonomous driving benchmarks, and their outputs serve as the historical context for prediction~\cite{shi2022motion, gao2023dynamic, wang2023prophnet, zhou2023query}. However, tracking remains brittle in crowded or occluded scenes, often producing fragmented trajectories or ID switches. In fact, most reported IDF1 \cite{chen2025study} scores are below 72\%, reflecting a substantial risk that can significantly degrade forecasting accuracy. Despite this limitation, nearly all forecasting methods still assume reliable tracking. To our knowledge, trajectory prediction in the complete absence of consistent IDs has not been explicitly studied. This gap is critical, as tracking failures are inevitable in practice and can persist long enough to jeopardize safety. Our work addresses this limitation by reconstructing historical trajectories directly from unlabeled detections, enabling reliable forecasting without stable tracking. Prior work has explored forecasting from detections without explicit track IDs, often relying on tracker intermediates such as affinity scores \cite{weng2022whose,zhang2023towards}. In this work, we explicitly study trajectory prediction under tracking failures without requiring any affinity estimation or other tracker outputs. 

\subsection{Scene Context Representation}
Scene context in motion forecasting refers to the complex traffic environment, including HD maps and multi-agent interactions, which autonomous vehicles must interpret to avoid collisions and conflicts~\cite{liu2023tracing, pellegrini2009you}. Early works rasterized the environment, storing maps and agent states as image-like tensors and applying convolutional networks for feature fusion~\cite{chai2019multipath, salzmann2020trajectron++}.  
With the introduction of VectorNet~\cite{gao2020vectornet}, scene context was unified into a vectorized representation of points, polylines, and polygons for both maps and trajectories. This paradigm has since become dominant~\cite{gu2021densetnt, zhou2022hivt, liu2024laformer, wang2023prophnet, gao2023dynamic, zhou2023query}, as vectorized representations provide broader receptive fields, finer map details, and lower computational cost. They further support encoding methods with permutation invariance and long-range dependency modeling, such as graph networks~\cite{scarselli2008graph} and Transformers~\cite{vaswani2017attention}, enabling direct interaction among agents and maps.  

Most forecasting methods normalize the scene to the target agent’s coordinates, aligning its heading with the $y$-axis (longitudinal direction)~\cite{zhou2022hivt, liu2024laformer, wang2023prophnet}. While effective, this approach requires re-normalizing at every timestep, which prevents reuse of encoded context. To address this limitation, QCNet~\cite{zhou2023query} introduced spatiotemporally invariant encodings, ensuring that agent and map features remain stable across timesteps and locations. This property not only supports efficient streaming prediction, but also provides the foundation for exploring trajectory forecasting under missing or unstable tracking, which we build upon in this work.

\subsection{Multimodal Trajectory Prediction}
Agent behavior is inherently uncertain, leading to multi-modal distributions of possible futures in complex environments. Multi-modal decoders are therefore critical for trajectory prediction. Generative models such as GANs~\cite{gupta2018social}, CVAEs~\cite{xu2022socialvae}, and diffusion models~\cite{xu2024controllable, yan2025trajflow} have been widely explored. Another line of work employs winner-takes-all strategies, where multiple prediction heads specialize in different motion modes~\cite{wang2023prophnet, zhou2022hivt, liu2024laformer}.  

Recent advances increasingly leverage Transformer decoders~\cite{vaswani2017attention}, inspired by DETR~\cite{carion2020end}, where each query directly decodes into a candidate trajectory, naturally supporting multi-modal forecasting~\cite{wang2023prophnet, gao2023dynamic, zhou2023query}. In this work, we adopt this query-based decoding paradigm, which effectively combines scene context with past trajectories to produce diverse and accurate future predictions.

\begin{figure*}[t!]
\centerline{\includegraphics[width=1\linewidth]{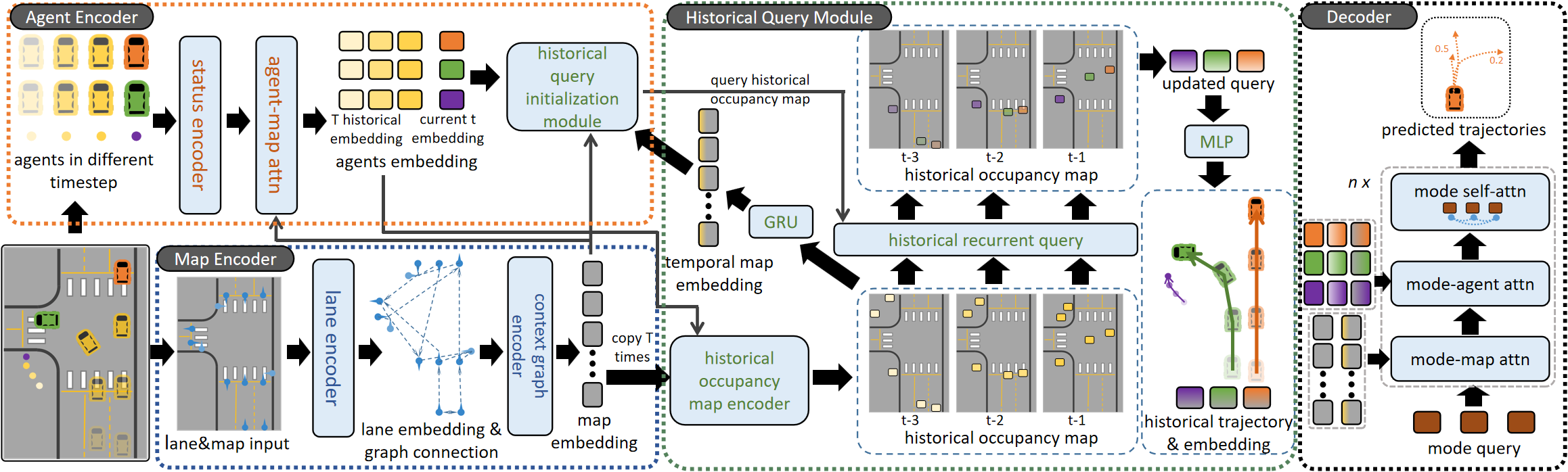}}
\caption{Overview of our \textbf{HiMAP} pipeline. 
    The framework consists of four main components: 
    \textbf{Agent and Map Encoding}, which embeds agent states and HD map elements in spatiotemporally invariant local frames; 
    \textbf{Historical Occupancy Map Encoder}, which aggregates per-frame agent--lane interactions into occupancy representations without relying on tracking IDs; 
    \textbf{Historical Query Module}, which initializes a history-aware query from the current agent state and iteratively attends to past occupancy maps to reconstruct agent-specific trajectories; 
    \textbf{Future Trajectory Decoder}, a DETR-style query decoder that generates multi-modal predictions from the reconstructed history, final query, and map context. 
    This design provides a robust fallback when tracking fails, enabling reliable forecasting directly from historical detections.}
\label{fig:2}
\vspace{-18pt}
\end{figure*}

\section{Method}
\label{Method}
\subsection{Problem Formulation}
We consider a scenario with $A$ agents surrounding the autonomous vehicle at discrete timesteps $t \in \{1,\dots,T\}$. During online operation, the perception stack provides to the predictor the states of detected agents at fixed intervals. When tracking is available, a data association module links detections across time to form past trajectories. For the $i$-th agent at time $t$, we denote its state feature as $\mathbf{s}_i^t = \{p_{i,x}^t, p_{i,y}^t, v_{i,x}^t, v_{i,y}^t, h_i^t, a_i^t\}$, where $p_i$ is the position, $v_i$ is the velocity, $h_i$ is the heading, and $a_i$ is the agent category (e.g., car, bus, pedestrian).
The HD map is represented by $M$ lane polygons, where the $j$-th polygon is $L_j = \{P_j, a_{L,j}\}$. Here $P_j = \{p_{j,k}\}$ is a set of sampled points with $p_{j,n} = \{p_{j,n,x}, p_{j,n,y}, a_{L,j,n}\}$ denoting position and lane point attributes, and $a_{L,j}$ encodes polygon-level attributes such as lane type or intersection membership. Given an observation of $T$ steps for an agent, the prediction module aims to generate $K$ possible future trajectories $\{Y_i^{(k)}\}_{k=1}^K$ along with their probabilities $\{\pi_i^{(k)}\}_{k=1}^K$.

In practice, however, tracking can be unstable or unavailable, leading to the loss of consistent agent identities over time. In such cases, the input within the observation window is no longer the identity-aware sequence $\{\mathbf{s}_i^t\}$ but instead a set of unlabeled detections per frame, denoted as $\mathcal{D}^t = \{\tilde{\mathbf{s}}_{n}^t\}_{n=1}^{N_t}$, where each $\tilde{\mathbf{s}}_{n}^t$ has the same structure as $\mathbf{s}_i^t$ but without a persistent agent ID, and there is no guaranteed correspondence across timesteps. The full observation history is then $\mathcal{D}_{1:T} = \{\mathcal{D}^t\}_{t=1}^T$. The prediction objective remains the same: given the current target agent and the HD map $\{L_j\}_{j=1}^M$, predict $K$ plausible future trajectories with probabilities, now conditioned on the identity-agnostic history $\mathcal{D}_{1:T}$ rather than tracked sequences. Our method is explicitly designed to handle both regimes by leveraging historical detection sets without relying on tracking IDs.

\subsection{Baseline Model}
We adopt QCNet~\cite{zhou2023query} as our baseline to investigate tracking failures. QCNet has two main advantages. First, it achieves excellent performance on the Argoverse~2 dataset, making it a strong backbone. Second, it employs spatiotemporally invariant encodings, ensuring consistent representations of agents and map segments across timesteps and locations.  

Unlike agent-centric methods that re-normalize scene context to the target agent’s position and heading at each frame, QCNet avoids redundant re-encoding. Such normalization prevents reuse of past context and requires ID-based trajectory storage, which is infeasible when tracking is unreliable.  
Because the map encoding space is invariant and agent encodings remain stable over time, historical states can be directly projected onto the map to construct spatiotemporally invariant occupancy information. Thus, QCNet supports efficient streaming prediction and joint forecasting of all agents under a unified, ID-agnostic representation, making it an ideal baseline for studying robustness to tracking failures.

\subsection{Overall Architecture}
Our prediction framework HiMAP consists of four modules shown in Fig.~\ref{fig:2}. 
The \textbf{Scene Encoding} module provides spatiotemporally invariant representations of agents and map elements. 
The \textbf{Historical Occupancy Map Encoder} aggregates per-frame agent--lane interactions into occupancy features without relying on tracking IDs. 
The \textbf{Historical Query Module} reconstructs agent-specific history by attending to occupancy maps conditioned on the current state. 
Finally, the \textbf{Future Trajectory Decoder} generates diverse multi-modal forecasts from the reconstructed history, the current state, and the map context. 
This modular design ensures robust prediction even under tracking failures.

\subsubsection{Scene Encoding}
We follow QCNet~\cite{zhou2023query}: agents and lanes are embedded in local frames for invariance. Lane polylines are encoded via cross-attention, while agent motion states are encoded with Fourier features and semantic attributes. Unlike QCNet, we do not use agent IDs: we store the set of visible agent embeddings at each timestep, forming $E_A \in \mathbb{R}^{T \times A_t \times D}$, together with lane embeddings $E_L \in \mathbb{R}^{M \times D}$, where $A_t$ denotes the number of visible agents at timestep $t$, $M$ the number of map elements, $T$ the observation length, and $D$ the feature dimension. For streaming inference, only the current frame is recomputed.

\subsubsection{Historical Occupancy Map Encoder}
We introduce a historical occupancy map encoder that allows the model to incorporate scene information from each past frame and capture long-term agent interactions. Since tracking data is unavailable, each frame contains an independent set of agent embeddings $E_A^t$. Leveraging the spatiotemporally invariant encodings, the connections between agents and lanes are represented as a graph, where each agent embedding is linked to nearby lane segments through relative positional encodings.  

For each historical frame, the occupancy map representation $E_\textrm{occ}^t$ is defined as
\begin{equation}
\label{OCC}
\begin{aligned}
E_\textrm{occ}^t &= E_L + \hat{g}^t \odot \Phi\big(E_A^t \rightarrow E_L,\, d(p_A^t \rightarrow p_L)\big), \\
\hat{g}^t &= \sigma\Big(\text{MLP}\big([E_A^t + d(p_A^t \rightarrow p_L)] \oplus E_L\big)\Big).
\end{aligned}
\end{equation}
where $E_L$ denotes lane embeddings, $d(p_A^t \rightarrow p_L)$ is the relative position encoding from agents to lanes, $\oplus$ denotes concatenation, and $\Phi(\cdot)$ is a cross-attention operator defined on the agent–lane graph. The gating vector $\hat{g}^t$ with Sigmoid activation $\sigma$ adaptively modulates the influence of agents on lane occupancy, reducing the contribution of agents with limited relevance to lane dynamics.  
In this way, each lane’s historical occupancy encoding is derived from both its own features and the gated contributions of nearby agents, conditioned on their relative positions. This design ensures that only behaviorally relevant agents exert a strong influence on the historical occupancy representation, thereby improving robustness to noisy or weak interactions. Collecting all timesteps, we obtain the historical occupancy tensor of shape $[T, M, D]$, where $T$ is the number of observed frames, $M$ is the number of lane segments.

\subsubsection{Historical Query Module}
To incorporate past occupancy information, we first summarize the sequence of historical occupancy maps $[T, M, D]$ into a temporal map embedding $\tilde{E}_L \in \mathbb{R}^{M \times D}$ using a GRU. This embedding compactly represents long-term map–agent interactions across $T$ frames.  

At the current timestep $t_c$, the observed agent state $E_A^{t_c}$ is connected to its surrounding map within an $r$-meter radius. Since map coordinates are spatiotemporally invariant, the relative position encoding $d(p_L \rightarrow p_A^{t_c})$ is shared across the temporal map embedding $\tilde{E}_L$ and all historical occupancy maps $E_\textrm{occ}^t$. The initial historical query is then formed by cross-attention between the current agent and the temporal map embedding:  
\begin{equation}
\label{eq:query_init}
A_\textrm{query}^{t_c} = E_A^{t_c} + \hat{g}^{t_c} \odot \Phi\big(\tilde{E}_L \rightarrow E_A^{t_c},\, d(p_L \rightarrow p_A^{t_c})\big),
\end{equation}
where $\hat{g}^{t_c} = \sigma\big(\text{MLP}([\tilde{E}_L + d(p_L \rightarrow p_A^{t_c})] \oplus E_A^{t_c})\big)$ is a gating vector, $\oplus$ denotes concatenation, and $\Phi(\cdot)$ is cross-attention on the agent–map graph. This design allows the query to integrate the current agent state with coarse localization and historical dynamics summarized by $\tilde{E}_L$.  

The historical query is then iteratively updated by attending to each occupancy map $E_\textrm{occ}^t$ in reverse temporal order, analogous to a recurrent update:  
\begin{equation}
\label{eq:query_iter}
A_\textrm{query}^{t-1} = A_\textrm{query}^t + \hat{g}^t \odot \Phi\big(E_\textrm{occ}^t \rightarrow A_\textrm{query}^t,\, d(p_L \rightarrow p_A^{t_c})\big),
\end{equation}
with $\hat{g}^t = \sigma\big(\text{MLP}([E_\textrm{occ}^t + d(p_L \rightarrow p_A^{t_c})] \oplus A_\textrm{query}^t)\big)$.  
Each updated query $A_\textrm{query}^{t-1}$ is projected by an MLP into a local displacement vector $h_{p}^{t-1}$, aligned with the coordinate frame of the current timestep. By iterating this process over $T_h$ historical steps with shared parameters, we obtain the reconstructed historical trajectory in local coordinates, analogous to an RNN unrolling through time. Importantly, our method queries each historical occupancy map conditioned on the current agent state, thereby implicitly retrieving the agent-specific information from unlabeled occupancy representations and decoding a coherent historical trajectory, which together with the final historical query $A_\textrm{query}^{T_h}$, the agent’s current state, and the map context, forms the input to our multi-modal trajectory decoder.

\subsubsection{Future Trajectory Decoder}
Built on the reconstructed history $h_{p}$, the final historical query $A_\textrm{query}^{T_h}$, and the map context $E_L$ together with the temporal map embedding $\tilde{E}_L$, we adopt a DETR-style decoder with $K$ learnable mode queries $\{q_k\}_{k=1}^{K}$. Each query with decoded historical trajectory attends to the scenario context and decodes one candidate future trajectory, yielding $K$ diverse hypotheses with associated probabilities:
\begin{equation}
\label{eq:traj_decoder}
\big\{\hat{Y}^{(k)},\, \pi^{(k)}\big\}_{k=1}^{K}
= \mathcal{D}\big(\{q_k\}_{k=1}^{K},\, A_\textrm{query}^{T_h},\, h_{p},\, \tilde{E}_L,\, E_L\big),
\end{equation}
where $\sum_{k=1}^{K}\pi^{(k)}=1$. This query-based design provides a compact, end-to-end mechanism for multi-modal forecasting while naturally leveraging the reconstructed history and current agent state.  

Unlike QCNet, which combines anchor-free and anchor-based decoding in a two-stage manner, we employ only the anchor-free design. The anchor-based refinement offers limited benefit in our setting, as our decoder already incorporates fine-grained historical features; our experiments show that it reduces FDE by only 0.03 while substantially increasing computational cost. Hence, the anchor-free stage alone is sufficient and more efficient for our framework, striking a better balance between accuracy and efficiency.

\begin{table*}[t!]
\begin{center}
\caption{Results on the Argoverse~2 motion forecasting benchmark. 
All leaderboard methods assume perfect tracking, whereas HiMAP is evaluated under the no-tracking setting (historical detections only). 
HiMAP achieves competitive results and clearly outperforms QCNet baselines when tracking fails.}
\vspace{-6pt}
\renewcommand{\arraystretch}{0.9}
\begin{tabular}{l|c|cc|cccc}
\toprule
\textbf{Method}&w/o Tracking&\textbf{minFDE$_1$}$\downarrow$ &\textbf{minADE$_1$}$\downarrow$&\textbf{minFDE$_6$}$\downarrow$ &\textbf{minADE$_6$}$\downarrow$&\textbf{MR$_6$}$\downarrow$&\textbf{b-minFDE$_6$}$\downarrow$\\
\midrule
FRM \cite{park2023leveraging} && 5.93 &  2.37 & 1.81 & 0.89 & 0.29 & 2.47\\
HDGT \cite{jia2023hdgt} && 5.37 &  2.08 & 1.60 & 0.84 & 0.21 & 2.24\\
MTR \cite{shi2022motion} && 4.39 &  1.74 & 1.44 & 0.73 & 0.15 & 1.98\\
HPTR \cite{zhang2023real}  && 4.61 &  1.84 & 1.43 & 0.73 & 0.19 & 2.03\\
HeteroGCN \cite{gao2023dynamic} && 4.40 &  1.72 & 1.34 & 0.69 & 0.18 & 1.90\\
ProphNet \cite{wang2023prophnet} && 4.74 &  1.80 & 1.33 & 0.68 & 0.18 & 1.88\\
QCNet \cite{zhou2023query} && 4.30 &  1.69 & 1.29 & 0.65 & 0.16 & 1.91\\
SmartRefine \cite{zhou2024smartrefine} && 4.17 &  1.65 & 1.23 & 0.63 & 0.15 & 1.86\\
DeMo \cite{zhang2024decoupling} && \textbf{3.74} &  \textbf{1.49} & \textbf{1.17} & \textbf{0.61} & \textbf{0.13} & \textbf{1.84}\\
\midrule
QCNet w/o tracking (baseline)& \checkmark & 11.59 & 5.88 & 3.23 & 2.00 & 0.52 & 3.89\\ 
DeMo w/o tracking& \checkmark & 9.99 & 5.14 & 3.16 & 2.09 & 0.47 & 3.85\\ 
QCNet w/ fine-tuning (baseline)& \checkmark & 5.16 & 2.10 & 1.49 & 0.77 & 0.21 & 2.25\\ 
HiMAP (Ours) & \checkmark & \textbf{4.61} & \textbf{1.81} & \textbf{1.33} & \textbf{0.68} & \textbf{0.17} & \textbf{1.96}\\
\bottomrule
\end{tabular}
\label{tab:av2}
\end{center}
\vspace{-18pt}
\end{table*}

\begin{table}[t!]
\begin{center}
\caption{Results on the Argoverse 1 motion forecasting validation set. }
\vspace{-6pt}
\renewcommand{\arraystretch}{0.9}
\begin{tabular}{l|ccc}
\toprule
\textbf{Method}&\textbf{minFDE$_6$}$\downarrow$ &\textbf{minADE$_6$}$\downarrow$&\textbf{MR$_6$}$\downarrow$\\
\midrule
QCNet\cite{zhou2023query} & 0.86 & 0.63 & 0.08 \\ 
\midrule
QCNet w/o tracking& 3.05 & 1.99 & 0.37 \\ 
QCNet w/ fine-tuning& 1.24 & 0.79 & 0.12 \\ 
HiMAP (Ours)& \textbf{0.94} & \textbf{0.66} & \textbf{0.09} \\
\bottomrule
\end{tabular}
\label{tab:av1}
\end{center}
\vspace{-18pt}
\end{table}

\subsection{Training}
Following prior work~\cite{zhou2022hivt, zhou2023query}, we model the distribution of predicted trajectories as a mixture of Laplace components. The regression loss is defined as the negative log-likelihood of the ground-truth trajectory under the closest mode $k_\textrm{best}$:  
\begin{equation}
\label{eq:Lreg}
 \mathcal{L}_\text{reg} = \frac{1}{t_f} \sum_{t=1}^{t_f} -\log P\big(Y_t \mid \mu_t^{k_\textrm{best}}, b_t^{k_\textrm{best}}\big),
\end{equation}
where $\mu_t$ and $b_t$ are the location and scale parameters of the Laplace distribution decoded by the trajectory predictor, $Y_t$ is the ground-truth position, and $t_f$ is the prediction horizon.  

To supervise the mode probabilities, we adopt a cross-entropy classification loss:  
\begin{equation}
\label{eq:Lcls}
 \mathcal{L}_\text{cls} = \sum_{k=1}^{K} -\pi_k \log(\hat{\pi}_k),
\end{equation}
where $\hat{\pi}_k$ is the predicted probability of mode $k$, $\pi_k$ is the one-hot ground-truth indicator, and $\sum_{k=1}^{K} \hat{\pi}_k = 1$.  

In addition, we directly supervise the reconstruction of historical trajectories decoded by the historical query module using an $\ell_2$ loss:  
\begin{equation}
\label{eq:Lhis}
 \mathcal{L}_\text{his} = \frac{1}{t_h} \sum_{t=1}^{t_h} \big\| \hat{h_p}_t^\text{his} - {h_p}_t^\text{his} \big\|_2^2,
\end{equation}
where $\hat{Y}_t^\text{his}$ and $Y_t^\text{his}$ denote the predicted and ground-truth historical positions, respectively, and $t_h$ is the number of historical steps.  

The final training objective is a weighted sum of the three terms:  
\begin{equation}
\label{eq:Lfinal}
 \mathcal{L} = \mathcal{L}_\text{reg} + \alpha \mathcal{L}_\text{cls} + \beta \mathcal{L}_\text{his},
\end{equation}
where $\alpha$ and $\beta$ are balancing coefficients.

\begin{table*}[t!]
\begin{center}
\caption{Ablation studies on Historical Query Module components. Experimental results are based on the Argoverse 2 validation set.}
\vspace{-6pt}
\setlength{\tabcolsep}{4pt}
\renewcommand{\arraystretch}{0.9}
\begin{tabular}{l|cccc|ccccc}
\toprule
\multirow{2}{*}{\textbf{Index}}&\multicolumn{4}{c|}{\textbf{Historical Query Module}} &\multicolumn{4}{c}{\textbf{Metrics}}\\
& recurrent query & hi. occ. map & hi. query init. & hi. temp. map &reconstruct hi. ADE$\downarrow$ &minFDE$_6$$\downarrow$ &minADE$_6$$\downarrow$&MR$_6$$\downarrow$&b-minFDE$_6$$\downarrow$\\
\midrule
1&  &  &  &  & ----- & 1.46 & 0.85 & 0.21 & 2.06 \\ 
2& \checkmark &  &  &  & 0.27 & 1.40 & 0.83 & 0.19 & 2.01 \\ 
3& \checkmark & \checkmark & &  & 0.21 & 1.38 & 0.79 & 0.18 & 2.00 \\ 
4& \checkmark & \checkmark & \checkmark &  & 0.14 & 1.37 & 0.79 & \textbf{0.17} & 1.99 \\ 
5& \checkmark & \checkmark & \checkmark & \checkmark & \textbf{0.12} & \textbf{1.33} & \textbf{0.76} & \textbf{0.17} & \textbf{1.96} \\ 
\bottomrule
\end{tabular}
\label{tab:ablation1}
\end{center}
\vspace{-6pt}
\end{table*}

\begin{table*}[t!]
\caption{Ablation studies on decoder inputs. Results are reported on the Argoverse 2 validation set.}
\vspace{-12pt}
\begin{center}
\renewcommand{\arraystretch}{0.9}
\begin{tabular}{l|cccc|cccc}
\toprule
\multirow{2}{*}{\textbf{Index}}&\multicolumn{4}{c|}{\textbf{Decoder}} &\multicolumn{4}{c}{\textbf{Metrics}}\\
& current agent status \& map  &  hi. temporal map & updated query & recurrent &minFDE$_6$$\downarrow$ &minADE$_6$$\downarrow$&MR$_6$$\downarrow$&b-minFDE$_6$$\downarrow$\\
\midrule
1& \checkmark &  &  & \checkmark & 1.46 & 0.85 & 0.21 & 2.06 \\ 
2& \checkmark & \checkmark &  & \checkmark &  1.39 & 0.79 & 0.18 & 2.01 \\ 
3& \checkmark & \checkmark & \checkmark &  & 1.41 & 0.83 & 0.19 & 2.03 \\ 
4& \checkmark & \checkmark & \checkmark & \checkmark & \textbf{1.33} & \textbf{0.76} & \textbf{0.17} & \textbf{1.96} \\ 
\bottomrule
\end{tabular}
\label{tab:ablation2}
\end{center}
\vspace{-18pt}
\end{table*}

\section{Experiments}
\subsection{Experimental Setup}
\subsubsection{Dataset}
We evaluate our model on the Argoverse~2 motion forecasting dataset~\cite{wilson2023argoverse}, which contains 25{,}000 real-world driving scenarios split into training, validation, and test sets with 199{,}908, 24{,}988, and 24{,}984 samples, respectively. Each sequence spans 11 seconds at 10~Hz, providing sufficiently long temporal horizons for both history and prediction. The large scale and extended duration of this dataset make it well-suited for assessing the temporal dependency and stability of our method.  

\subsubsection{Metrics}
We adopt standard evaluation metrics~\cite{gu2021densetnt}: $minADE_K$, $minFDE_K$, $b$-$minFDE_K$, and $MR_K$. $minADE_K$ measures the average displacement error between the ground-truth trajectory and the closest of $K$ predicted trajectories across all timesteps. $minFDE_K$ evaluates the final displacement error at the prediction horizon. $b$-$minFDE_K$ further incorporates the confidence score of each predicted trajectory. $MR_K$ denotes the miss rate, defined as the proportion of cases where the final displacement error exceeds 2 meters.  

\subsubsection{Implementation Details}
We train our model on 8 NVIDIA RTX~3090 GPUs using the AdamW optimizer~\cite{loshchilov2017decoupled}. The training schedule consists of 64 epochs with a batch size of 16 and an initial learning rate of $1 \times 10^{-4}$. We adopt a warm-up phase of 10{,}000 steps, followed by a cosine annealing scheduler~\cite{loshchilov2016sgdr} to decay the learning rate. The number of heads in all multi-head attention layers is set to 8. No ensemble methods or data augmentation are applied in our experiments.

\subsection{Comparison on Argoverse~2 Benchmark}
We evaluate our proposed \textbf{HiMAP} on the Argoverse~2 motion forecasting benchmark~\cite{wilson2023argoverse}. For fair comparison, ensemble methods are excluded. All leaderboard entries assume perfectly accurate tracking, whereas our method is designed as a fallback solution under tracking failures, relying only on historical detections without identity information. As shown in Table \ref{tab:av2}, despite this stricter setting, HiMAP still outperforms most existing forecasting approaches on the leaderboard.  

To further validate the practicality of our method, we investigate the performance of QCNet~\cite{zhou2023query} under tracking failures. Using the publicly released checkpoint and masking out tracking IDs, QCNet’s performance drops sharply due to distributional shift: $minFDE_6$ increases from 1.29 to 3.23, $minADE_6$ from 0.65 to 2.00, and $MR_6$ from 16\% to 52\%, rendering the predictions nearly unusable for autonomous driving. We then fine-tune the QCNet decoder under the no-tracking setting by randomly masking past agent trajectories, which partially recovers performance but still yields inferior results ($minFDE_6=1.49$, $minADE_6=0.77$, $MR_6=21\%$).

By contrast, HiMAP achieves $minFDE_6=1.33$, $minADE_6=0.68$, and $MR_6=17\%$, corresponding to relative improvements of 11\% in FDE, 12\% in ADE, and 4\% in MR over the fine-tuned QCNet baseline. These gains highlight the robustness of our approach in handling tracking failures, effectively bridging the gap between tracking-based forecasting methods and realistic perception scenarios. Overall, HiMAP demonstrates both strong predictive performance and robustness to tracking failures, highlighting its potential as a reliable module for safety-critical autonomous driving.

\subsection{Results on Argoverse~1 Dataset}
We further evaluate our proposed method on the Argoverse~1 validation set for comparison with our baseline in Table \ref{tab:av1}. Following the settings in QCNet~\cite{zhou2023query}, we reproduce its results on Argoverse~1 and benchmark our tracking-free variant under the same conditions. The results show that our method significantly improves performance in the absence of tracking and achieves accuracy close to tracking-based models, thereby ensuring safer and more reliable trajectory forecasting for autonomous driving.

\begin{figure}[t]
\centering
\includegraphics[width=0.75\linewidth]{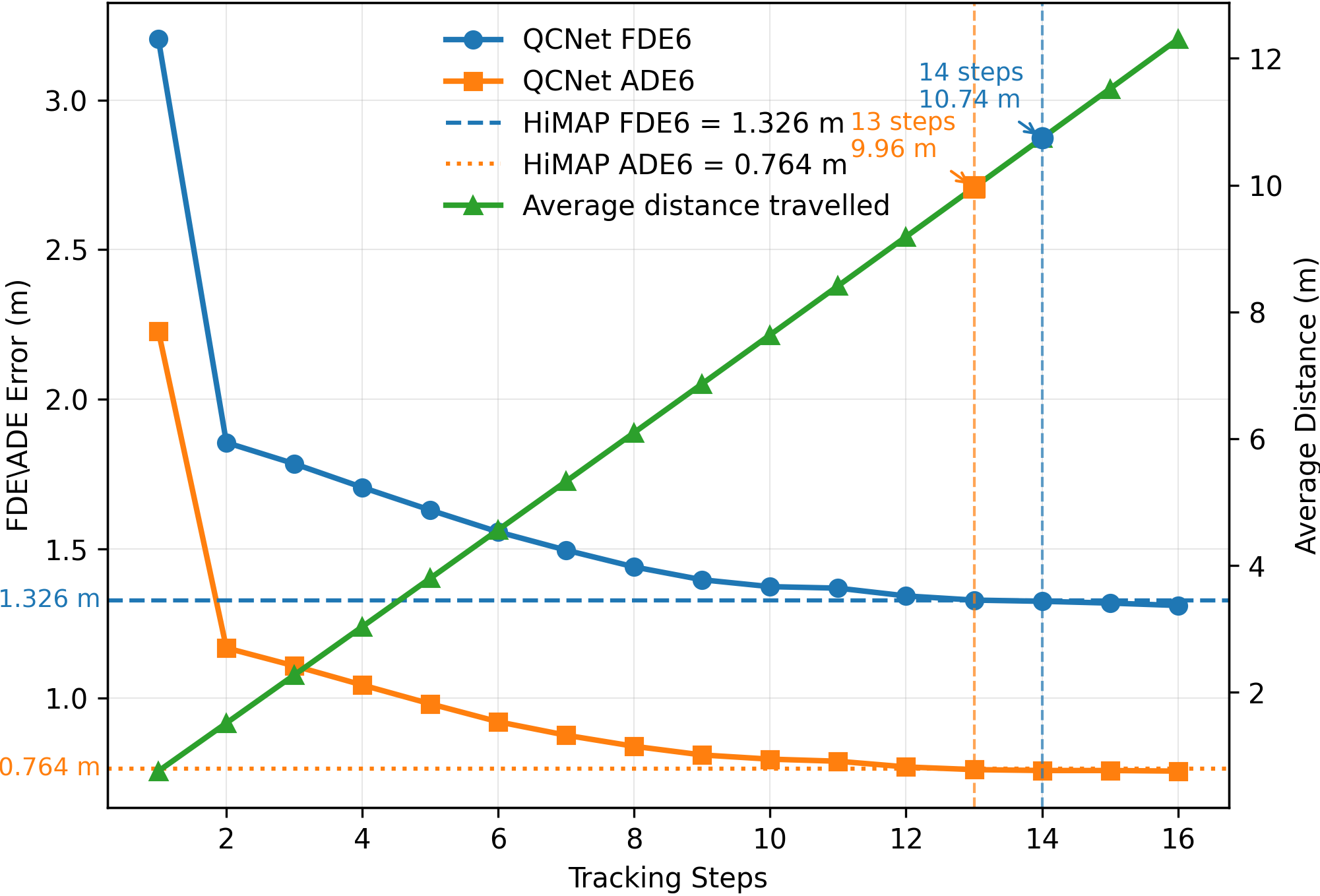}
\caption{Comparison between HiMAP and QCNet on the Argoverse 2 validation set under different numbers of available tracking steps. 
HiMAP maintains fixed performance without requiring tracking, shown as horizontal dashed lines. 
QCNet gradually improves as more tracked history is available, surpassing HiMAP after 13--14 steps. 
The green curve indicates the average distance traveled per timestep. }
\label{fig:p3}
\vspace{-18pt}
\end{figure}

\subsection{Impact of Tracking Failures}
Figure~\ref{fig:p3} illustrates the necessity of our study and its implications for autonomous driving safety. The plot compares the performance of HiMAP with QCNet on the Argoverse 2 validation set under varying numbers of available tracking steps.  

HiMAP achieves fixed performance on this benchmark ($minFDE_6=1.33$\,m and $minADE_6=0.76$\,m), shown as the horizontal dashed lines. In contrast, the curves show how QCNet’s errors decrease as more historical tracking steps become available. The green curve represents the average distance traveled by vehicles per timestep in the validation set, which increases approximately linearly.  

The figure highlights a critical safety gap. When tracking fails, QCNet requires about 13 steps (1.3\,s) for ADE and 14 steps (1.4\,s) for FDE to surpass HiMAP’s fixed accuracy. During this recovery period, vehicles travel on average 10.7\,m and fast-moving agents can cover over 30\,m, before QCNet regains its predictive reliability. Such delays pose substantial risks for real-time decision-making in safety-critical systems.  
By contrast, HiMAP immediately provides stable predictions without relying on tracked histories, improving short-term safety. Within the first second, HiMAP reduces FDE by 20\% and ADE by 23\% compared to QCNet without tracking on average, ensuring reliable trajectory forecasts precisely when they are most critical. This underscores the importance of designing tracking-agnostic forecasting modules. This robustness entails a trade-off between accuracy and efficiency, as each additional history-reconstruction step increases average latency by $1.6\%$.

\begin{figure*}[t!]
\centerline{\includegraphics[width=1\linewidth]{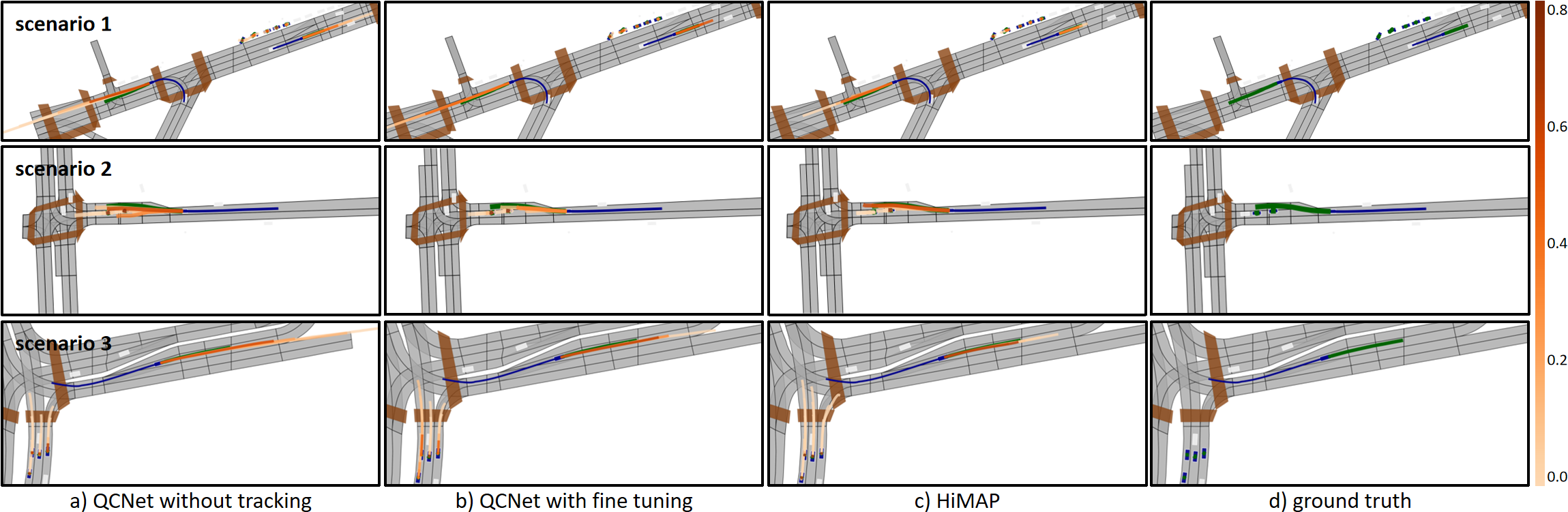}}
\caption{Qualitative comparison of trajectory prediction results on the Argoverse 2 validation set. 
The blue box denotes the target agent. Blue lines show the agent’s past trajectory (not used during prediction), green lines represent the ground-truth future trajectory, and orange lines indicate predicted trajectories, with darker colors corresponding to higher predicted probabilities, and the colorbar is on the right.}
\vspace{-18pt}
\label{fig:4}
\end{figure*}

\subsection{Ablation Study}
We conduct ablation studies to analyze the contribution of each component in the proposed Historical Query Module and Decoder, with results reported in Table~\ref{tab:ablation1} and Table~\ref{tab:ablation2}.  

Table~\ref{tab:ablation1} evaluates the Historical Query Module. The first row removes the entire module, which is equivalent to fine-tuned QCNet. In this case, prediction relies only on the current agent state and map, failing to reconstruct historical trajectories and yielding poor results. The second row allows the current agent state to directly query past agent states without occupancy maps. This design becomes infeasible, as the computational complexity at each timestep scales exponentially with the number of agents, and the query vector is easily entangled with unrelated agents across timesteps, leading to degraded reconstruction. Moreover, when the observed history of an agent is shorter than the required reconstruction horizon, this approach introduces substantial noise. In contrast, using occupancy maps enables the model to infer unobserved timesteps from map structure, yielding more plausible historical trajectories. The third row introduces historical occupancy maps, which substantially improve performance by providing a natural filtering mechanism and topological constraints, reducing complexity and improving retrieval accuracy. The fourth row adds historical query initialization, fusing the current agent state with map features before querying to refine trajectory recovery. Finally, incorporating a temporal map via GRU in the fifth row achieves the best results, as it models temporal dynamics in occupancy maps and queries, allowing the network to capture both evolving scene dynamics and agent-specific history.

Table~\ref{tab:ablation2} reports the ablation on decoder inputs. The first row uses only the current state and map, again equivalent to QCNet, and performs poorly in the no-tracking setting. The second row adds the historical temporal map, yielding large gains as decoded history is injected into DETR-style prediction queries (Section~\ref{Method}). The third row introduces the updated query, where agents interact with historical map states, enriching the temporal context. The fourth row adds recurrence~\cite{zhou2023query, zhou2024smartrefine}, improving long-horizon accuracy. The complete design in the fourth row achieves the best performance, showing that updated queries and recurrent decoding are both crucial for modeling agent-specific dynamics.

\subsection{Historical Timesteps Study}
\begin{table}[t!]
\begin{center}
\caption{Ablation studies on reconstructed historical steps $T_h$. Results are reported on the Argoverse 2 validation set.}
\vspace{-6pt}
\renewcommand{\arraystretch}{0.9}
\begin{tabular}{c|ccc}
\toprule
\textbf{reconstructed historical steps $T_h$} &\textbf{minFDE$_6$$\downarrow$} &\textbf{minADE$_6$$\downarrow$}\\
\midrule
10 & 1.37 & 0.78 \\ 
20 & 1.34 & \textbf{0.76} \\ 
30 & \textbf{1.33} & \textbf{0.76}\\ 
40 & 1.35 & 0.77 \\ 
50 & 1.36 & 0.77 \\ 
\bottomrule
\end{tabular}
\label{tab:ablation3}
\end{center}
\vspace{-18pt}
\end{table}
We conduct experiments on the number of reconstructed historical timesteps to determine the most suitable setting. Table~\ref{tab:ablation3} shows the impact of different choices. Using 30 timesteps achieves the best performance. Fewer timesteps may not provide sufficient historical context for accurate forecasting, while too many introduce noise and instability. In the Argoverse 2 dataset, many agents have fewer than 30 valid past observations, reflecting the practical challenge that real-world vehicles often lack long historical trajectories, which can further increase prediction noise.

\subsection{Qualitative Results}
Figure~\ref{fig:4} presents qualitative results comparing the baselines with our approach. Our method serves as a complementary fallback when tracking is unavailable, effectively reconstructing historical trajectories to ensure safer prediction. This reconstruction enables more accurate adjustment of agent speed and improved awareness of surrounding agents. In contrast, the original QCNet without tracking relies only on the current velocity, producing six map-constrained but overly simplistic predictions. After fine-tuning, QCNet can partially adapt its speed to scene context, yet its perception of motion dynamics and interactions remains limited. By incorporating reconstructed historical features, our approach provides robust and reliable forecasting even under tracking failure, thereby strengthening safety guarantees for autonomous driving.

\section{Conclusion}
In this work, we introduced HiMAP, a trajectory prediction framework designed to operate robustly under tracking failures. Unlike existing approaches that rely on stable tracking identities, HiMAP reconstructs historical trajectories from unlabeled detections by leveraging spatiotemporally invariant occupancy maps and a historical query mechanism. Our design enables accurate and reliable forecasting even when tracking modules fail.  
Experiments on Argoverse~2 show that HiMAP attains performance close to state-of-the-art tracking-based methods and substantially outperforms baselines in the no-tracking setting. These results highlight the importance of forecasting under imperfect perception and demonstrate HiMAP as a practical step toward safer and more reliable autonomy.

\bibliography{icra}
\end{document}